\documentclass[conference]{IEEEtran}
\IEEEoverridecommandlockouts
\usepackage{cite}
\usepackage{amsmath,amssymb,amsfonts}
\usepackage{algorithmic}
\usepackage{graphicx}
\usepackage{textcomp}
\usepackage{xcolor}
\usepackage{mdframed}

\usepackage{times}
\usepackage{latexsym}
\usepackage{graphicx}
\usepackage{lscape}
\usepackage{comment}
\usepackage{caption}
\usepackage{subcaption}
\usepackage{float}
\usepackage{url}
\usepackage[section]{placeins}
\usepackage{adjustbox}

\def\BibTeX{{\rm B\kern-.05em{\sc i\kern-.025em b}\kern-.08em
    T\kern-.1667em\lower.7ex\hbox{E}\kern-.125emX}}
\begin{document}

\title{Text2Time: Transformer-based Article Time Period Prediction\\}

\author{\IEEEauthorblockN{ Karthick Prasad Gunasekaran}
\IEEEauthorblockA{ \textit{University of Massachusetts}\\
Amherst, USA \\
kgunasekaran@umass.edu}
\and
\IEEEauthorblockN{B Chase Babrich}
\IEEEauthorblockA{ \textit{University of Massachusetts}\\
Amherst, USA \\
bbabrich@umass.edu}
\and
\IEEEauthorblockN{Saurabh Shirodkar}
\IEEEauthorblockA{ \textit{University of Massachusetts}\\
Amherst, USA \\
sshirodkar@umass.edu}
\and
\IEEEauthorblockN{Hee Hwang}
\IEEEauthorblockA{ \textit{University of Massachusetts}\\
Amherst, USA \\
hhwang@umass.edu}
}

\maketitle

\begin{abstract}
The task of predicting the publication period of text documents, such as news articles, is an important but less studied problem in the field of natural language processing. Predicting the year of a news article can be useful in various contexts, such as historical research, sentiment analysis, and media monitoring. In this work, we investigate the problem of predicting the publication period of a text document, specifically a news article, based on its textual content. In order to do so, we created our own extensive labeled dataset of over 350,000 news articles published by The New York Times over six decades. In our approach, we use a pretrained BERT model fine-tuned for the task of text classification, specifically for time period prediction. This model exceeds our expectations and provides some very impressive results in terms of accurately classifying news articles into their respective publication decades. The results beat the performance of the baseline model for this relatively unexplored task of time prediction from text.
\end{abstract}

\section{Introduction}

The problem of determining the time period in which a text document, such as a news article, was published is crucial, yet it has been less explored in the domain of natural language processing. The problem has a wide range of use cases that can be applied to different domains. For example, in historical research, predicting the year of a news article can assist scholars in understanding the context and significance of events from a particular time period, as it aims to understand past events and their impact on society, culture, and politics. To do so, scholars rely heavily on primary sources such as newspapers, journals, and historical documents. Nevertheless, a significant number of historical articles lack a publication date, which makes it challenging for scholars to conduct research. 

The problem can also be applied in the field of sentiment analysis, where predicting the year of a news article can provide insights into how public opinion has evolved over time. For example, imagine a sentiment analyst examining public opinion on climate change over the past decade. By predicting the year of a news article, the analyst can track how sentiment towards climate change has evolved over the years. They may discover that public opinion was largely indifferent to climate change in the early 2010s but shifted towards greater concern in the mid-2010s. This information could be used to inform public policy decisions or marketing strategies. Another example of the use case is in the field of media monitoring, where predicting the year of a news article can aid in tracking trends and identifying patterns in media coverage. It can also be useful in verifying the authenticity of old news articles and preventing the spread of misinformation.

Whether or not the kind of language used in a text document can give us information about the publication date of the document is a very interesting question. Google Ngram Viewer is a fantastic resource that provides us with a lot of insights about this question. For example, a quick visual analysis on Google Ngram Viewer shows that the name `Sherlock Holmes' started appearing in English text documents in 1890, and got progressively more and more frequent until 1930. It then got progressively less frequent until 1970, following which it regained its popularity and has been frequently used since. Such trends of word and phrase usage certainly seem to give some indication of the publication period of a text.

Given a text document, the problem we have chosen is to estimate the time period during which the document was written. Identifying whether or not there exist linguistic properties in text documents such as news articles, that are indicative of the time period in which the text was written, seems to be an interesting problem with high impact in the field of computational linguistics.

We start by collecting our own labelled dataset by using the New York Times news articles API. The dataset is fairly extensive with over 350,000 news articles published over six decades, and requires a significant amount of preprocessing and cleaning. The details of this process are mentioned in Sections 1.2, 1.3, 1.4 and 1.5.

Next, we create a simple baseline text classification model for predicting the publication decade of a news article from its text. For the baseline we choose Naive Bayes, and still manage to achieve an impressive accuracy of 63\% using this fairly simplistic model. The details of the baseline model are mentioned in Section 4.

Finally, for our best model, we use a state of the art pretrained BERT model, and fine tune it for the task of multiclass text classification. This model achieves a very impressive and surprising accuracy of 82\% on our test data. These results far outperform any prior results from previously tried models \cite{Bakarji} for this task of prediction of publication period from text. The architecture, implementation details and results achieved by our BERT model are described in detail in section 5.

Our major contributions in the paper include:

\begin{itemize}
\item {A novel dataset of new articles}
\item {Our BERT based time period prediction model outperforms current baseline approaches}
\item {We perform in-depth qualitative and quantitative error analysis to discuss the examples our approach struggles with}

\end{itemize}

\section{Literature Review}

Our work was inspired by the idea of generating headlines of news articles from their text. Much work has already been done in this area with many different models, ranging from dependency trees ~\cite{LiqunWang} to RNNs ~\cite{Lopyrev2015} to viewing the problem as a machine translation issue ~\cite{BankoMittal}. We suspect that the features generated for this problem will be similar to the ones generated for ours.

Date prediction of news articles is a much more sparse field of work. We could only find one paper which explored the exact problem we had come up with ~\cite{Bakarji}. Their work applied three different kinds of models: Bayesian, logistic regression, and neural networks. In each of these cases their models are fairly simple and we hope to distinguish our work by applying more complex ones. 

Temporal prediction in general is a slightly more explored problem, but previous papers focus on historical texts rather than contemporary ones ~\cite{deJong}. Therefore there is much more data available for our problem. However, the methods used in their models may still be useful for us to consider, such as producing rankings of dates rather than specific dates ~\cite{Niculae}.

\section{DATASET}

\subsection{Building Dataset}

All the pre-existing datesets of news articles we could find only contained articles dating back to around the year 2000, when digital articles started becoming much more freely available. We hypothesized that using data over a larger span of time that that, i.e., around 100 years, would allow for more linguistic variation and therefore yield more interesting results. Therefore, we decided to use the New York Times Archive API in combination with a Python library build to scrape newspaper articles to scrape over the last 100 years and build our dataset \cite{newyorktimesapi} \cite{newspaper3k}. Ultimately our data ended up taking the form of individual raw data files for each article as well as meta file which contained the data file's ID, year, month, and category.

\begin{figure}[!htb]
    \centering
           \begin{mdframed}

    \textit{8,The Islanders moved to within one game of clinching their Stanley 
    Cup semifinal playoff last night dominating the Rangers in every aspect of play and beating them 5-1 at Madison Square Garden.The victory meant...} 
        \end{mdframed}

    \caption{A single data example. The 8 at the beginning denotes that the article came from the 1980s}
    \label{fig:my_label}
\end{figure}

\subsection{Cleaning Dataset}

Some documents in the dataset had some default phrases by New York Times included along with the article text itself. Following are some examples of such phrases.

\begin{figure}[!htb]

    \centering
        \begin{mdframed}

    \textit{
    \begin{itemize}
    \item Credit... The New York Times Archives \\
    \item Full text is unavailable for this digitized archive article. Subscribers may view the full text of this article in its original form through TimesMachine. \\
    \item NYTimes.com no longer supports Internet Explorer 9 or earlier. Please upgrade your browser.
    \end{itemize}}
    \end{mdframed}

    \caption{Default phrases added by NYTimes}
    \label{fig:my_label}
\end{figure}

We detected and removed these phrases from each article as they would not give our models a true representation of the article text to predict the date from, causing bias in the prediction process.

Additionally, a bigger issue was that some articles had the publication date itself included with the article text in the format \textit{Month Day, Year}. Having these dates present in the articles while performing a publication time classification from the text would mean that the classifier had direct access to the publication date and could easily cheat. We use regular expressions to detect the presence of such a publication date and remove it from each article.

Figures 4 and 5 show an example of an article in the dataset before and after the cleaning process.
\begin{figure}[!htb]
    \centering
    \begin{mdframed}

    \includegraphics[scale=1.0]{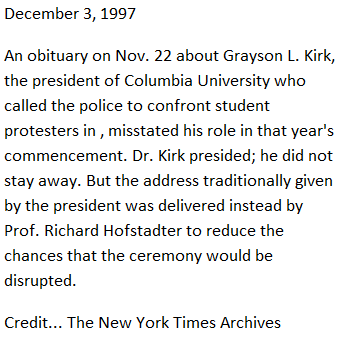}
\end{mdframed}

    \caption{Article before cleaning}
    \label{fig:my_label}
\end{figure}

\begin{figure}[!htb]
    \centering
    \begin{mdframed}

    \includegraphics[scale=1.0]{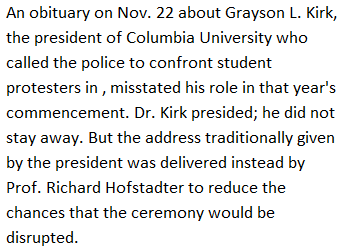}
\end{mdframed}

    \caption{Article after cleaning}
    \label{fig:my_label}
\end{figure}

\subsection{Filtering Dataset}

In order to better study the data and understand what our models were learning from it, we decided to stratify our data based on the category it was written under. Over the course of 60 years, around 250 different category types had been used. Therefore, we grouped each subcategory into the following larger categories by hand. Here is a summary of our data stratification based on article category:

\begin{figure}[!htb]
    \centering
    \begin{tabular}{ |c | c| }
     \hline
     Category & no. articles \\
     \hline\hline
     Art, Fashion, Food and Wine & 12,922 \\
     \hline
     Blogs, OpEds, and Obituaries & 19,649 \\  
     \hline
     Books and Magazines & 7,096\\
     \hline
     Business and Finance & 49,910\\
     \hline
     Domestic and Culture & 53,623\\
     \hline
     International & 26,778\\
     \hline
     Lifestyle & 7,206\\
     \hline
     News & 2,615\\
     \hline
     Science And Tech & 3,308\\
     \hline
     Entertainment & 1,498\\
     \hline
     Sports & 17,136\\
     \hline
    \end{tabular}
    \caption{Number of articles in each category}
    \label{fig:my_label}
\end{figure}

We admit that doing the splitting by had incurs a certain amount of bias and so that might in part explain why there is not a more even distribution of articles across categories. It is also important to note that the sum of the right column does not equal the total number of articles we scraped due to the fact that there is also an omitted row for the miscellaneous category. We have omitted that row because any linguistic variation found in that category would be due to the randomness of the genres of writing and not due to trends in a single genre of writing.

In addition to splitting the data up by categories, we also produced filtered versions of the dataset using these various criteria:

\begin{itemize}
  \item Articles truncated to average article length in training data
  \item Data with all years within actual article text removed
  \item Data split up by decade instead of categories
\end{itemize}

These were created to yield either statistics about the dataset or properties about our models.

\subsection{Dataset Statistics}

Our algorithm to build our dataset was as simple as using \cite{newyorktimesapi} to grab the URLs of each article and then using \cite{newspaper3k} to scrape the article's text. However, even though the algorithm was simple it was still limited to scraping one article at a time and so scraping articles was a relatively slow process. Indeed, the downloading was split across multiple machines and additional project work had to be put into writing scripts to merge large non-contiguous chunks of data.

Due to this fact, as well as the fact that digitized versions of the articles become more and more sparse as we go further and further back in time, we only ended up scraping data from the year 1960 until now. We still managed to scrape 367,809 articles, which is around 61,301 articles per decade, more than enough to yield reasonable results on our models. Here are some more statistics about our dataset:

\begin{figure}[!htb]
    \centering
    \begin{adjustbox}{width=0.5\textwidth}
    \begin{tabular}{ |c|c|c|c|c|c|}
     \hline
     Decade & Min & Max & Mean & Median & Std Dev \\
     \hline\hline
     1960s & 13 & 1314 & 51.8 & 35.0 & 46.5 \\
     \hline
     1970s & 7 & 1959 & 197.2 & 165.0 & 119.5 \\
     \hline
     1980s & 5 & 8361 & 196.0 & 164.0 & 152.6\\
     \hline
     1990s & 3 & 3673 & 216.9 & 187.0 & 140.2\\
     \hline
     2000s & 4 & 9964 & 302.7 & 200.0 & 372.7\\
     \hline
     2010s & 7 & 17284 & 317.1 & 214.0 & 396.0\\
     \hline
    \end{tabular}
    \end{adjustbox}
    \caption{Per Decade Article Length Statistics}
    \label{fig:my_label}
\end{figure}

\begin{figure}[!htb]
    \centering
    \includegraphics[scale=0.5]{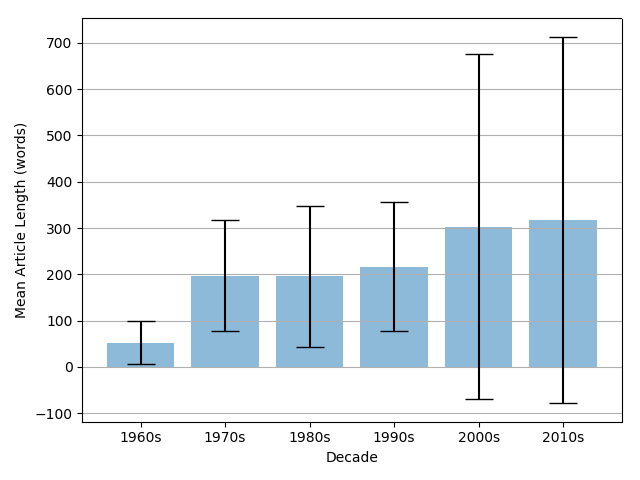}
    \caption{Per Decade Mean Article Length with Std Dev Bars}
    \label{fig:my_label}
\end{figure}

\begin{figure}[!htb]
    \centering
    \includegraphics[scale=0.5]{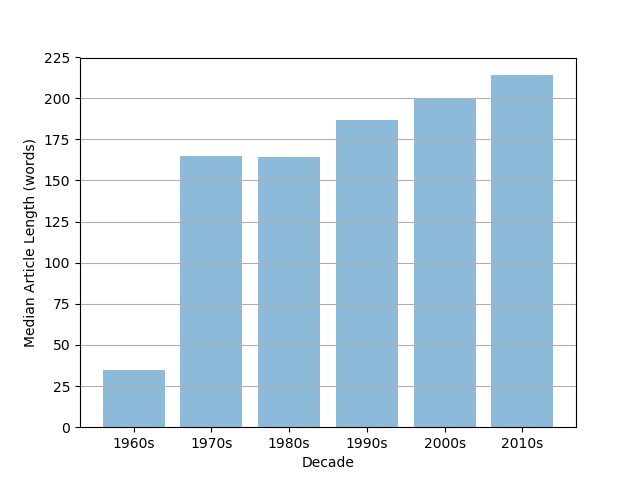}
    \caption{Per Decade Median Article Length}
    \label{fig:my_label}
\end{figure}

Figure 7 shows the minimum, maximum, mean, median and standard deviation of the word lengths for all articles in every decade that we have considered.

Figure 8 shows a plot of the mean and standard deviation of the article word lengths for every decade, and Figure 9 shows the median for the same.

We can observe that the average and median word length of articles is increasing with each decade, being the lowest in 1960s and highest in 2010s. Additionally, the standard deviation of word lengths also has increased significantly in latter decades. It is possible that our BERT model picked up on this phenomenon of word length of the article being correlated with the decade in which it was published, and it might be one of the reasons for the impressive performance that it achieves.

\section{Baseline}

For our baseline we simply used Naive Bayes to predict which decade an article was written given its text. On our full and cleaned dataset, with 74,093 articles to train on and 29,317 articles to test on, this baseline model achieved an accuracy of 63\%. 

The brighter color indicates that the decade got a high accuracy. For example, the 1970s is the class that receives a high number of correct articles. The heatmap shows bright colors diagonally, which explains that the ground-truth and predicted labels are aligned.

We also believed the year in the article could affect our results. After taking out all year information, the accuracy of Naive Bayes becomes 61\%.
\begin{figure}[!htb]
    \centering
    \includegraphics[scale=0.35]{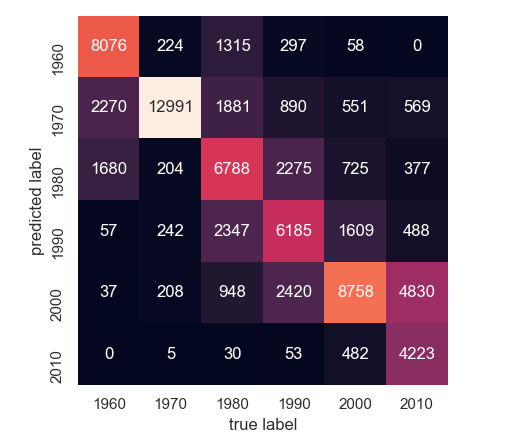}
    \caption{Heatmap for Baseline (Naive Bayes) On Full and Cleaned Dataset}
    \label{fig:my_label}
\end{figure}

\begin{figure}[!htb]
    \centering
    \includegraphics[width=0.5\textwidth]{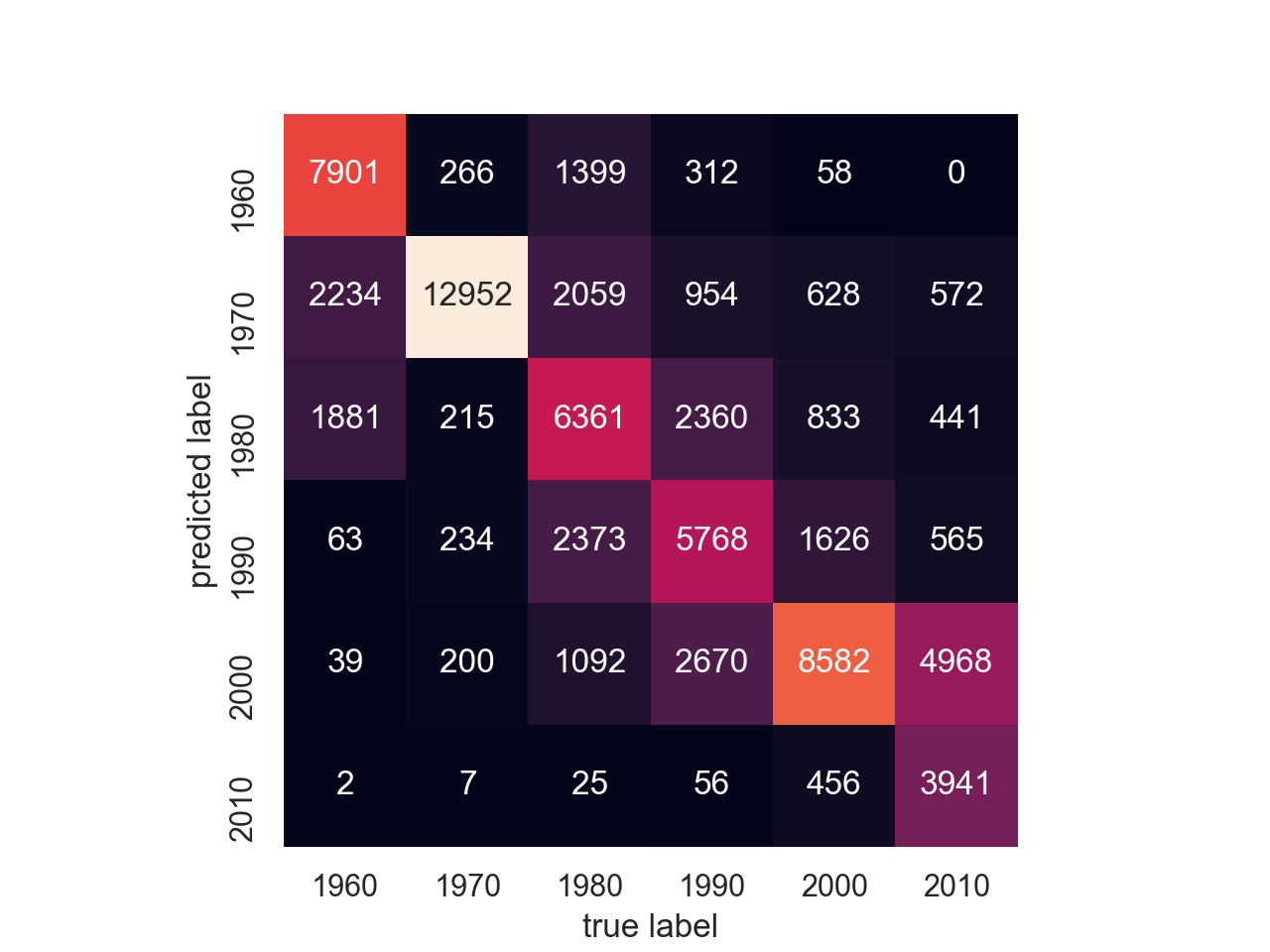}
    \caption{Heatmap for Baseline (Naive Bayes) On Dataset without year information}
    \label{fig:my_label}
\end{figure}

\newpage
\begin{figure}[!htb]
    \centering
    \includegraphics[width=0.5\textwidth]{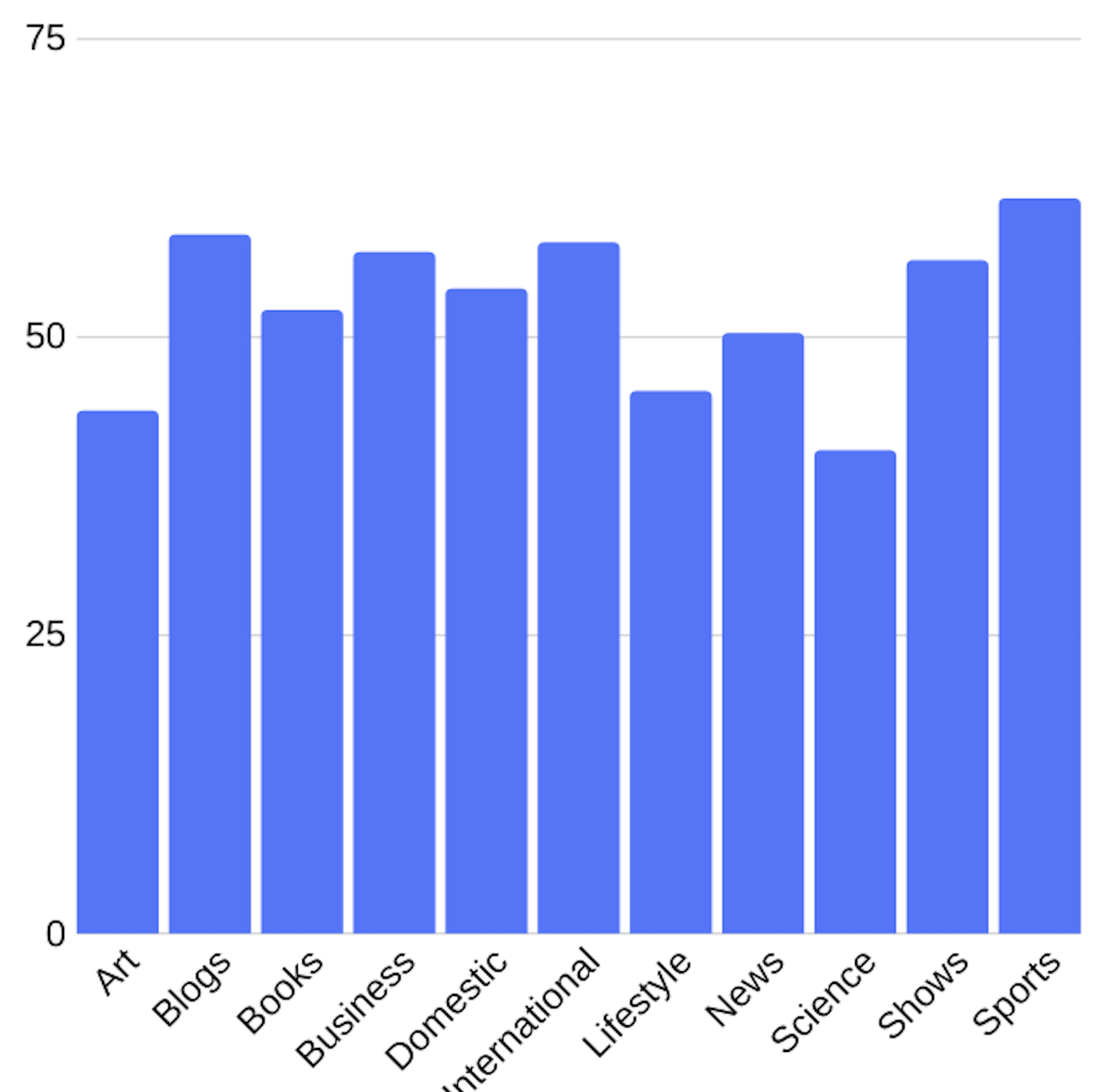}
    \caption{Naive Bayes Accuracies by categories}
    \label{fig:my_label}
\end{figure}

We trained the model by categories as well. It does not achieve high accuracies; however, we could see some interesting findings, which are:
\begin{itemize}
\item Lack of data($\leq$ 3000 articles) could explain News, Science, Autos, and Tech's poor accuracies. 
\item However, the Books and Magazine category gave us good accuracy, although it has only 5675 articles.
\item The Lifestyle category is tough to determine its decade; In other, we could say that lifestyle does not change that much from the 1960s to 2010s.
\end{itemize}
\section{Our Approach}

\subsection{Architecture}

After producing the raw version of our data set, which was just individual files for every article and an overarching meta file for referencing, we had to convert each file to .csv format. These .csv files were then once more transformed to be in the special "tsv" format that the huggingface libraries prefer \cite{huggingfaces}. Following these transformations, the data was finally ready to be turned into embeddings.

\begin{figure}[!htb]
    \centering
    \includegraphics[scale=0.4]{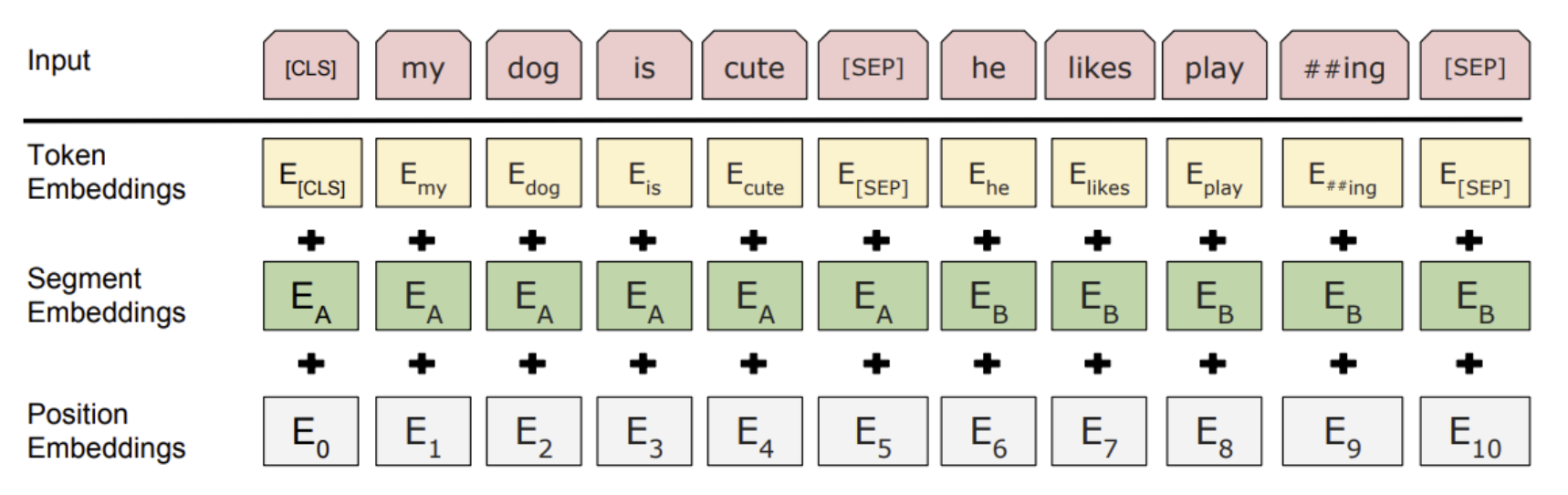}
    \caption{Transforming textual inputs into embeddings for Bert}
    \label{fig:my_label}
\end{figure}

After all of that, we are finally able to feed our data as input into a pre-trained BERT model. Here are the details of the model we used:

\begin{itemize}
    \item Pre-trained  Bert-Base cased model
    \item 12-layer, 768-hidden, 12-heads, 110M parameters
    \item Connected layer, Softmax classifier used for fine tuning
\end{itemize}

As implied by the last item on the above list, the parameters in our BERT model remained frozen during our training. All of the learning that was done took place in the last layer of our architecture, and that was a feed-forward neural network with five fully connected layers and a softmax classifier placed at the end for predicting decades.

\begin{figure}[!htb]
    \centering
    \includegraphics[scale=0.5]{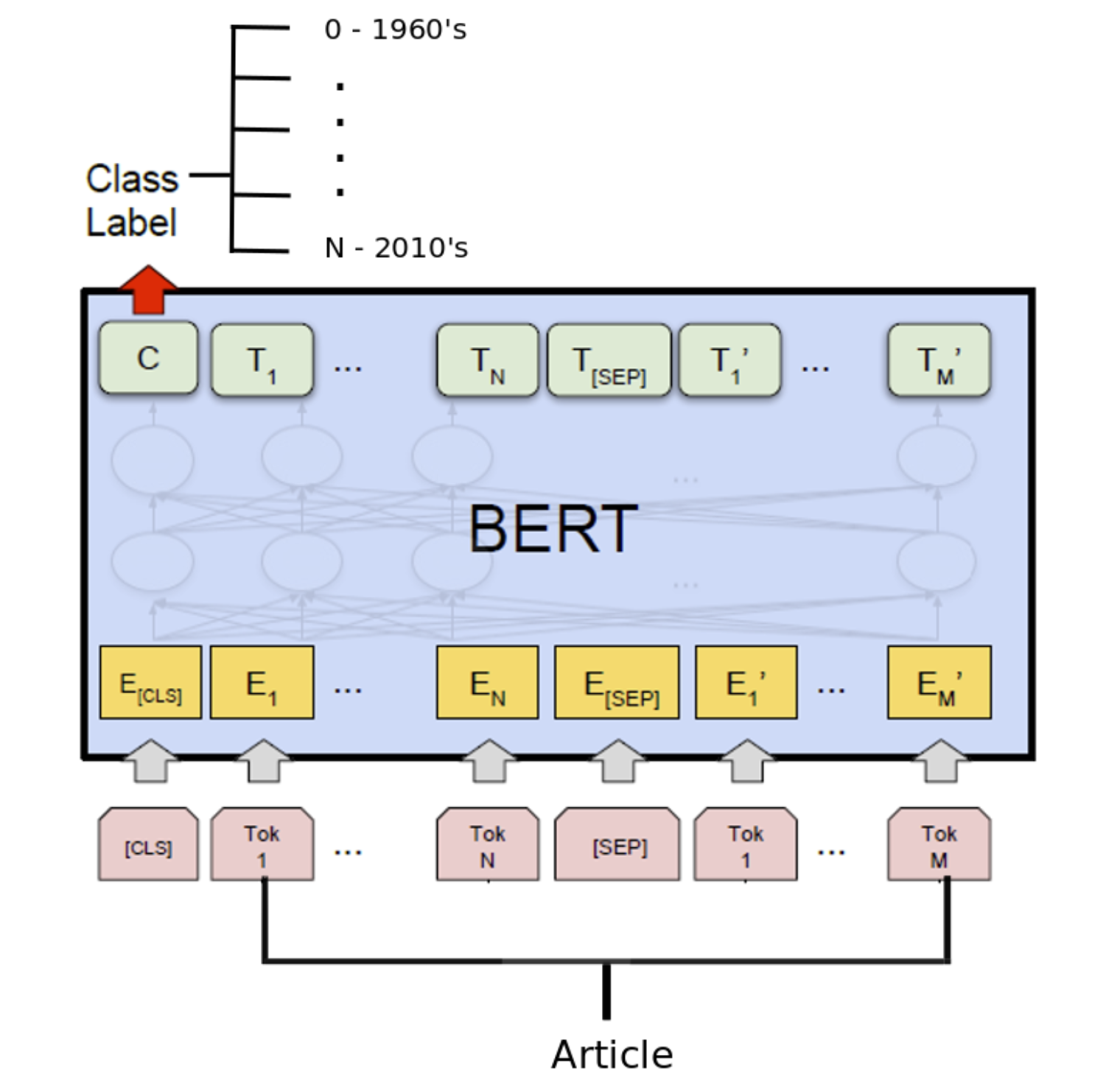}
    \caption{Architecture of our BERT Model}
    \label{fig:my_label}
\end{figure}

\subsection{Implementation}

In order to implement BERT, we used a library which is built on top of the huggingface library ~\cite{huggingfaces}. We trained our dataset on pretrained baseline version of Bert. A cased bert was used since the classification would be depending on cased character which might important compared to non-cased character. Also, since earlier articles around 1960s didn't have proper fullstops signalling sentence endings we felt using pretraianed cased BERT would be a better option than using without it. The BERT which was used was the default one with 12-layers which consists of 768 hidden units and heads. The BERT has approximately around 110M parameters on top of it.

The entire training took approximately 10 1/2 hours of time per run running on GPU in Google Colaboratory \cite{colab}. Training batch size of 16 and testing batch size of 8 articles were used. The learning rate used was 4e-5 which was ideal since decreasing it led to slow gradient updates and increasing it led to unusual and irregular gradient updates. Adam optimizer was used while training and regularization parameter was set to 1e-8. Training was restricted to 2 epochs due to the limiting factor only 12 hours of continuous GPU usage in Google Colab. Also, it was noticed that gradient updates seemed small after 2 epoch so extra steps weren't taken for running for more epochs.

After training, the model was able to achieve an accuracy of 82 percent on the test data. To prove the BERT results are ideal Various analysis have been presented on the following sections and results have been tabulated.

\subsection{Loss Metric}
    Since this a multi class classification problem softmax function where cross entropy loss was used to train and evaluate the model.Following equation was used for training and evaluation purpose 
    \begin{equation}
        L_i = -\lg(\frac{exp^{f_{y_i}+ \lg C}}{\sum_j exp^{f_j + \lg C} })
    \end{equation}
    
    where $f_{y_i}$ are the scores for each class and C is a small constant value for numerical stability.
\subsection{Results}

\begin{table*}[ht]
\centering
 \caption{\label{tab:dataset_table} Model Accuracy across different datasets}
 
\begin{tabular}{|l |c|c|c|c|c|c|}
\hline
 
\textbf{Dataset} & \textbf{1960s} & \textbf{1970s}& \textbf{1980s} & \textbf{1990s} & \textbf{2000s} & \textbf{2010s} \\ \hline
\textbf{Original dataset} & 93.67 & 94.81 & 85.16 & 71.58 & 59.85 & 89.26         \\ \hline
\textbf{Years Removed} & 93.07 & 94.20 & 83.15& 68.57& 58.29& 85.13 
\\ \hline
\textbf{Uniform Length} & 91.86 & 93.77 & 82.23 & 65.48 & 58.28 & 89.39 
\\ \hline

\end{tabular}
\end{table*}

     Bert's performance was evaluated on the downstream date classification task and compared with the naive bayes approach. The classification task was performed with six classes or six decades ranging from 1960s to 2010s. Our BERT classifier was able to achieve an accuracy of 82\% on classification using entire dataset. However, there were some doubts on the model performance on whether the article content  years data could have resulted in this high accuracy or the length of the articles might have inflated the accuracy.So various tests where performed which have been explained below to prove that the classifier was not predicting based on the article length or the years in article data. Also, comparison was performed across different categories of articles to examine if some types of articles have high relevance in predictions.     
          
  \begin{figure}[!htb]
   \includegraphics[width=0.5\textwidth]{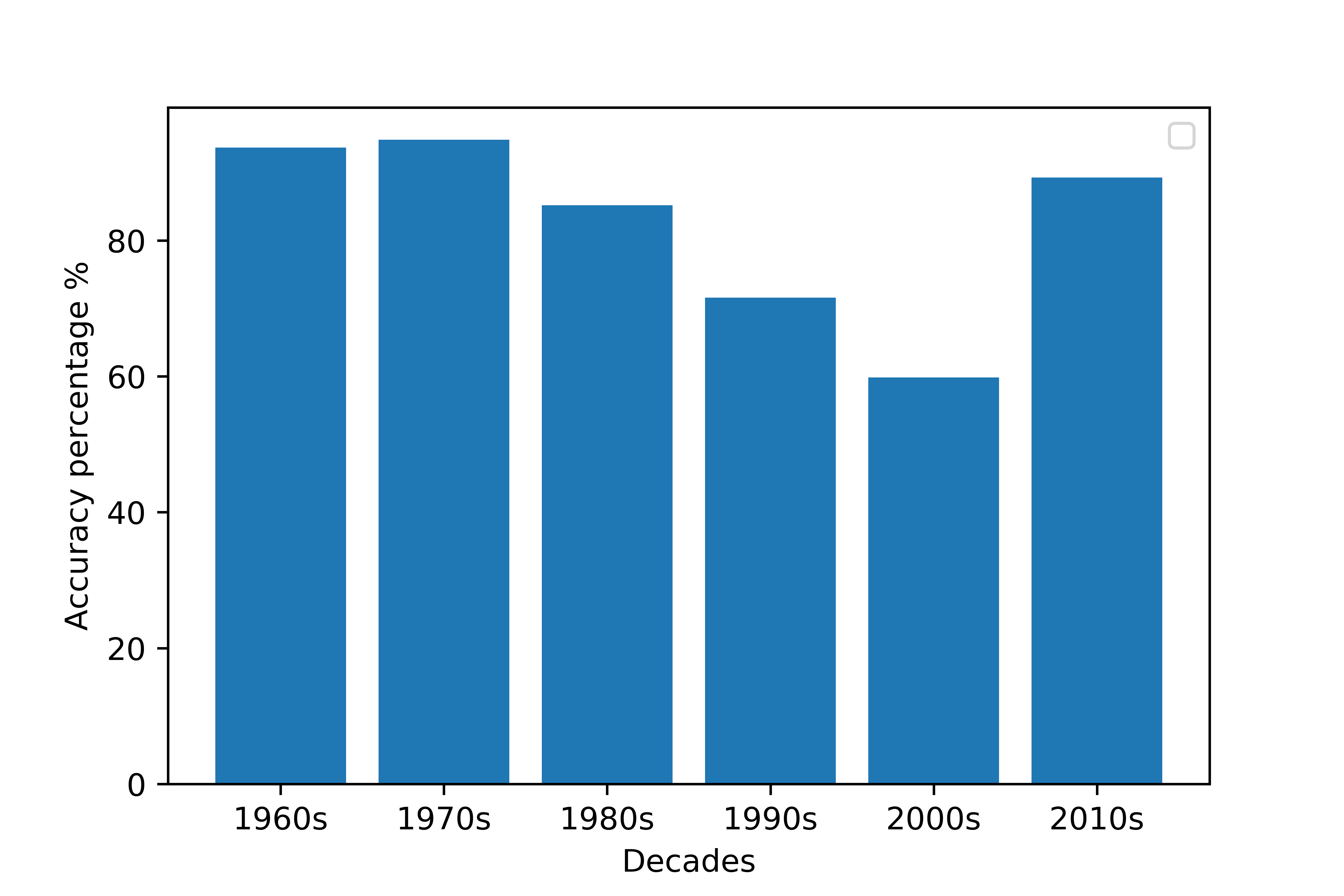}
    \caption{Results of BERT classifier}
    \label{fig:original_plot}
\end{figure}
 
 \begin{figure}[!htb]
   \includegraphics[width=0.5\textwidth]{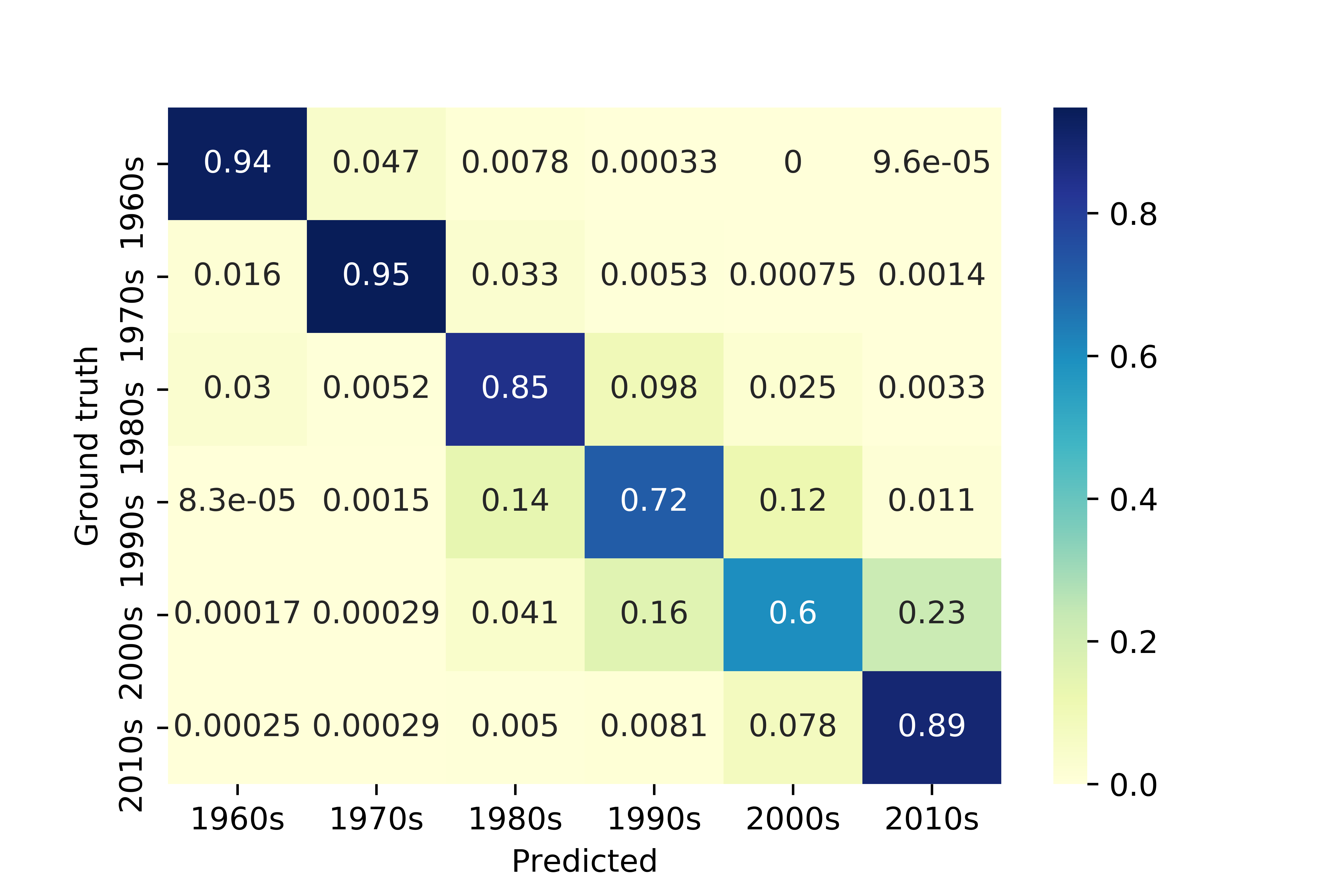}
    \caption{Correlation matrix for BERT results}
    \label{fig:original_map}
\end{figure}
     From the Figure \ref{fig:original_plot} it can be seen the classifier predicts the articles belonging be 1960s, 1970s and 2010s more accurately than other three. Further from the correlation map shown in Figure \ref{fig:original_map}. The predictions that are wrong are mostly from nearby decades. So from this it can be observed that the articles have distinct common nouns, writing styles, and content. In Figure \ref{fig:original_map} it can be seen that the 2000s decade is mostly wrongly predicted as 2010s and 1990s. One reasonable answer to this can be due to the fact that nouns might be popular across decades which might trick the classifier into predicting the wrong decade.

 \begin{figure}[!htb]
   \includegraphics[width=0.5\textwidth]{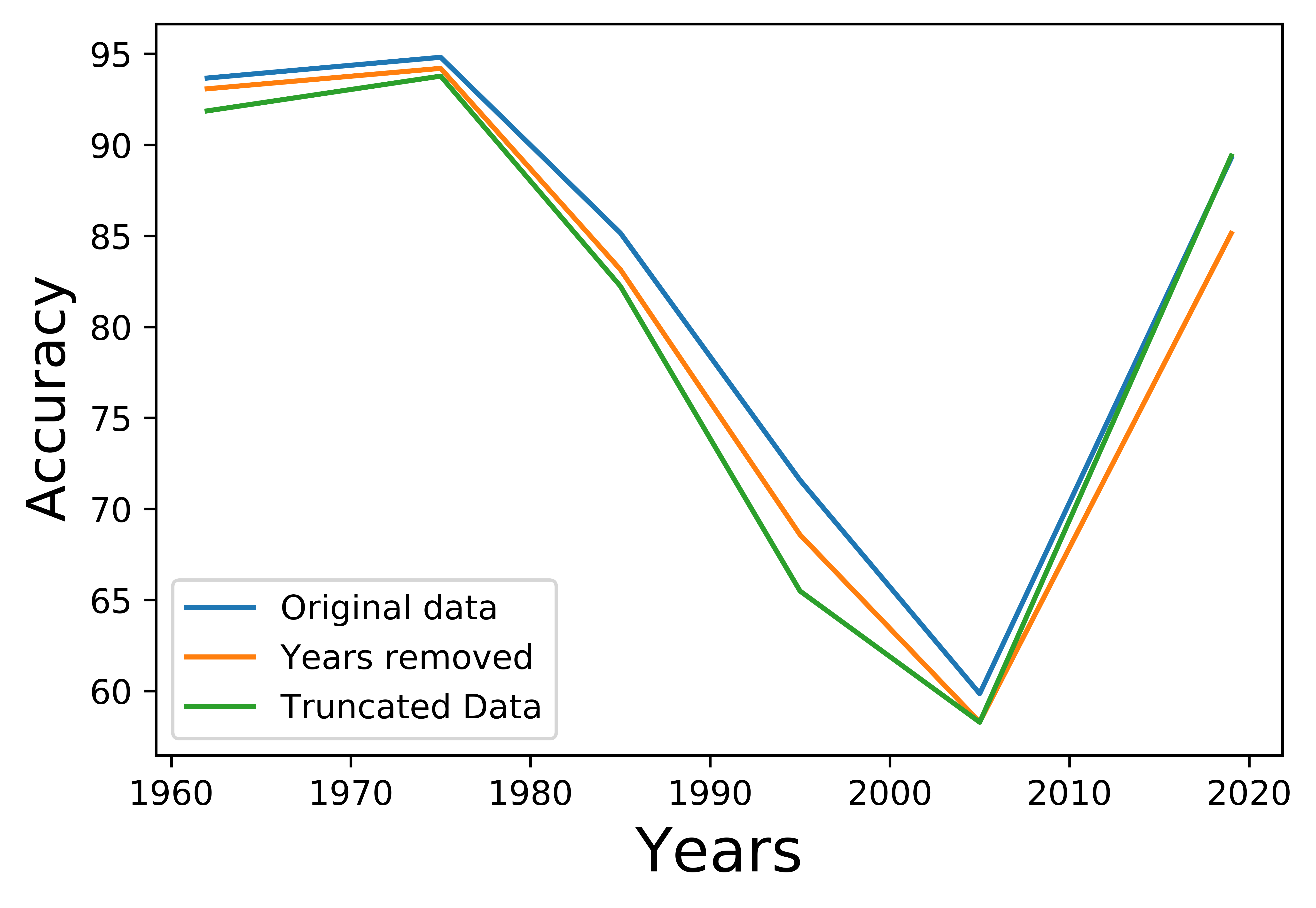}
    \caption{Effect of dates and uniform Article length}
    \label{fig:dates_length_original}
\end{figure}

   \subsubsection{Comparison with Baseline Approach}
   A comparison of the BERT results with the Naive Bayes baseline model used was made.  BERT classifier outperforms the naive bayes classifier in all the categories and across all the decades.  Table \ref{tab:naive_bayes_years} shows the comparison of the naive bayes vs the BERT classifier. From the table it can clearly be seen that the BERT accuracy outperforms the accuracy of naive bayes in all decades expect the 2000s.

\begin{table}[H]

 \caption{\label{tab:naive_bayes_years}Model Accuracy Comparison by Decade}
\begin{tabular}{|l|l|l|}
\hline
\textbf{Decade} & \textbf{Naive Bayes} & \textbf{BERT} \\ \hline
1960s           & 66.63                & 93.67         \\ \hline
1970s           & 93.63                & 94.81         \\ \hline
1980s           & 51.00                & 85.16         \\ \hline
1990s           & 51.03                & 71.57         \\ \hline
2000s           & 72.88                & 59.85         \\ \hline
2010s           & 40.26                & 89.26         \\ \hline
\end{tabular}
\end{table}

Table \ref{tab:category_table} shows the accuracy of the naive bayes classifier and  BERT classifier across all the categories. It can be seen that the model almost beats the baseline by at least 10\% across all the categories. This shows that the BERT is able to handle the context much better than the Naive Bayes. It should be noted that the fine tuning is 10x more expensive than performing naive Bayes classification in terms of time and computational resources. However, With these results we are able to conclude that BERT model outperforms techniques classical machine learning techniques such as Naive Bayes.

\begin{table}[]
 \caption{\label{tab:category_table} Model Accuracy Comparison by Category}
 
\begin{tabular}{|l|l|l|}
\hline
 
\textbf{Category}            & \textbf{Naive Bayes} & \textbf{BERT} \\ \hline
Art, Fashion, Food, and Wine & 43.80                & 68.19         \\ \hline
Blogs                        & 58.54                & 80.08         \\ \hline
Books and Magazines          & 52.24                & 66.43         \\ \hline
Business and Finance         & 57.08                & 80.35         \\ \hline
Domestic and Culture         & 54.02                & 78.29         \\ \hline
International                & 57.90                & 80.46         \\ \hline
Lifestyle                    & 45.41                & 66.50         \\ \hline
News                         & 50.28                & 71.48         \\ \hline
Science and Tech             & 40.44                & 63.70         \\ \hline
Shows, Movies, Games         & 56.39                & 72.13         \\ \hline
Sports                       & 61.60                & 74.99         \\ \hline
\end{tabular}
\end{table}

\subsubsection{Effect of removing dates in the article}

\begin{figure}[!htb]
 \centering
     \begin{subfigure}[b]{0.44\textwidth}
           \centering
         \includegraphics[width=\textwidth]{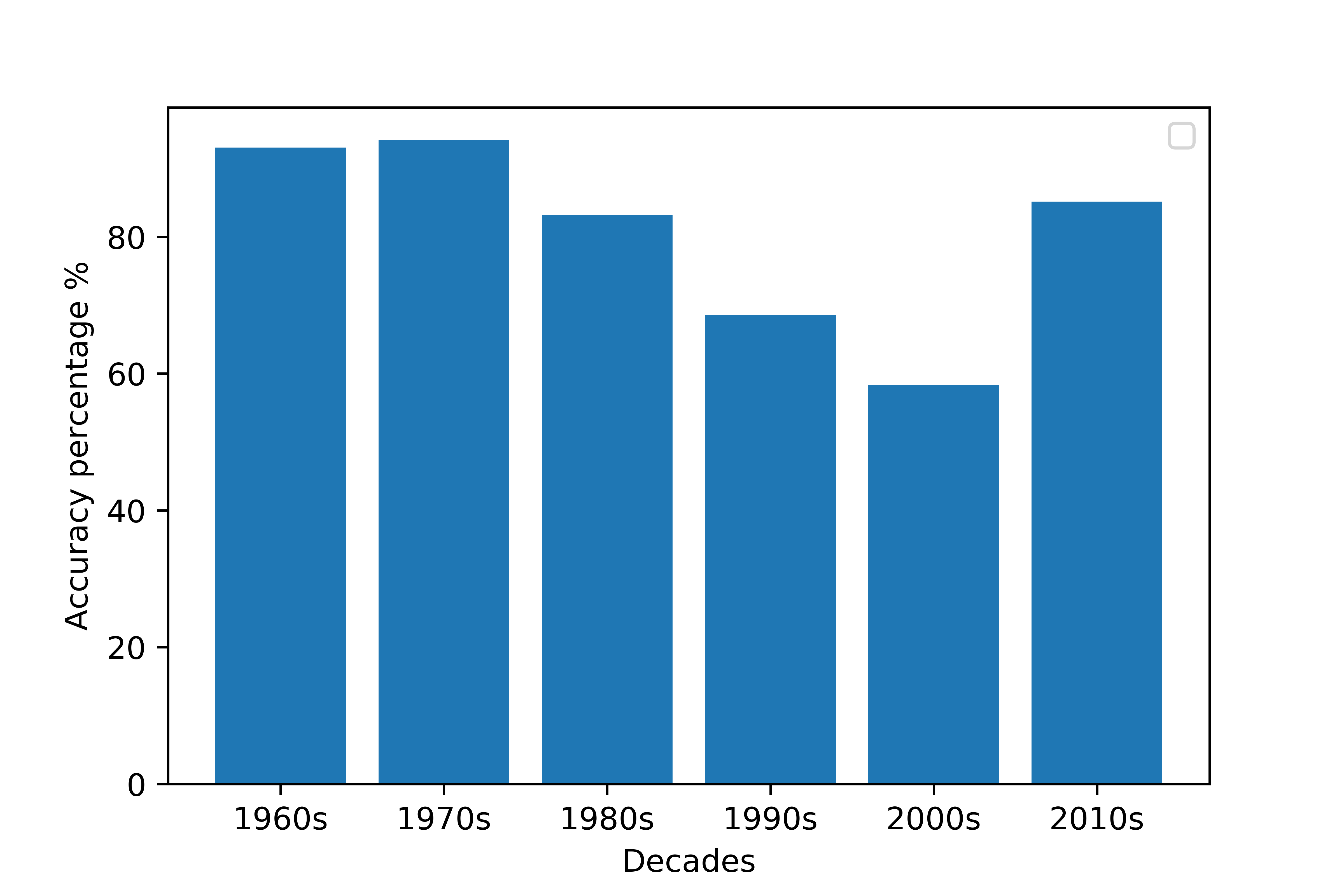}
       \caption{Plot showing accuracy decade wise post removing dates in articles}
         \label{fig:dates_plot}
     \end{subfigure}
     \centering
     \begin{subfigure}[b]{0.44\textwidth}
           \centering
         \includegraphics[width=\textwidth]{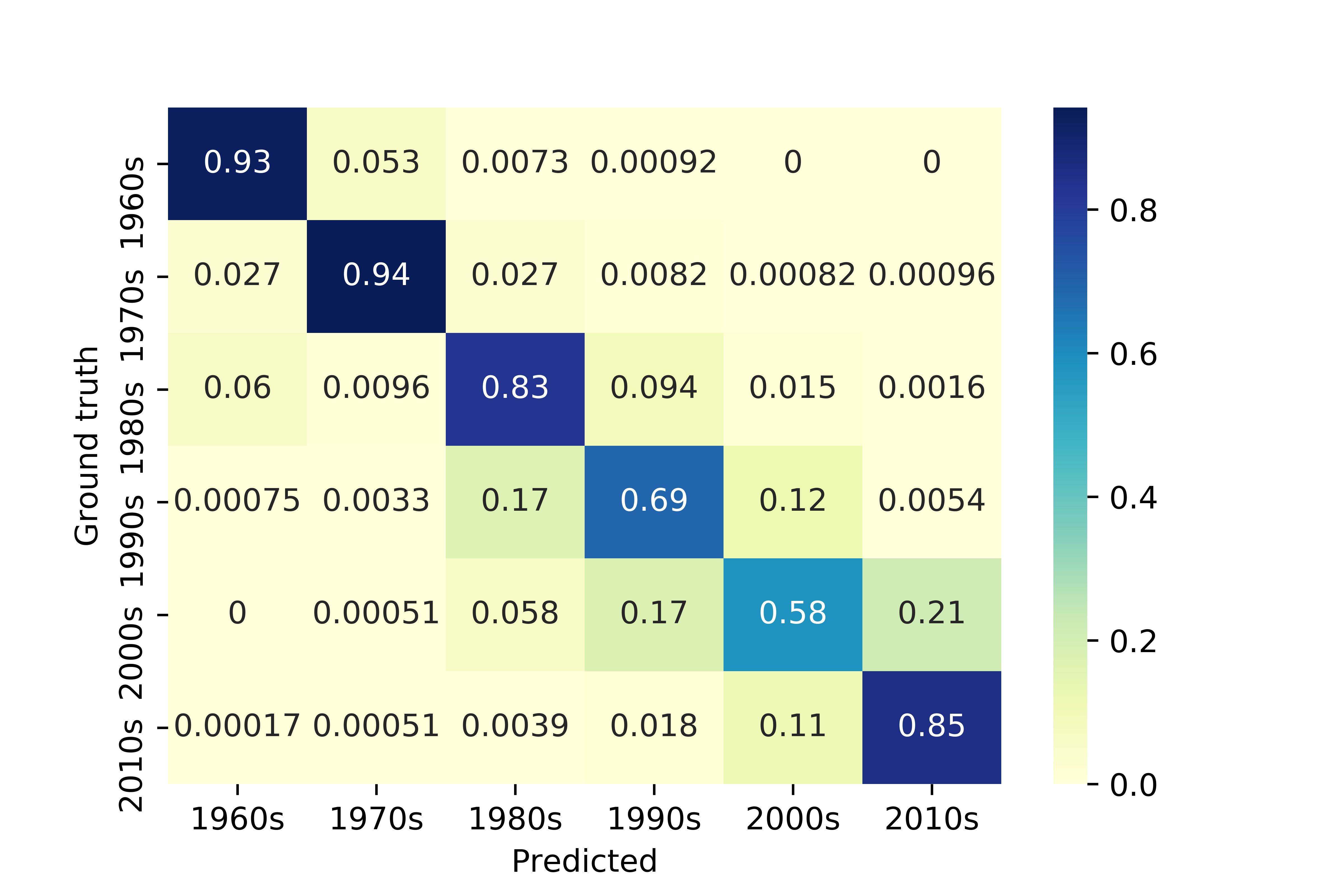}
     \caption{Correlation matrix after dates removed}
         \label{fig:dates_map}
     \end{subfigure}
         \caption{Plots showing the results after dates removed}
    \label{fig:dates}
\end{figure}

 To prove that there is no effect of years in the data set. Dataset was processed to remove all the years in the articles. The dates were not limited to 1960-2019, but all the dates from 1800s were removed. This was done since there might be articles in particular decade which refers frequently to particular year range. The model trained and tested on this dataset has an accuracy of around 79\% which almost similar to the original dataset.  Figure \ref{fig:dates_length_original} plotted shows that the accuracy pattern is very much similar to the pattern seen in original data. Figure \ref{fig:dates_plot} shows the decade-wise accuracy for the data set with dates removed. Also, almost same correlation pattern is seen in Figure \ref{fig:dates_map} as the original data. Years and dates in the dataset are essential and natural features of the articles; however, we have shown that there is not too much relevance on the dates in the dataset.

\subsubsection{Effect of Article Length}
\begin{figure}[!htb]
 \centering
     \begin{subfigure}[b]{0.44\textwidth}
           \centering
         \includegraphics[width=\textwidth]{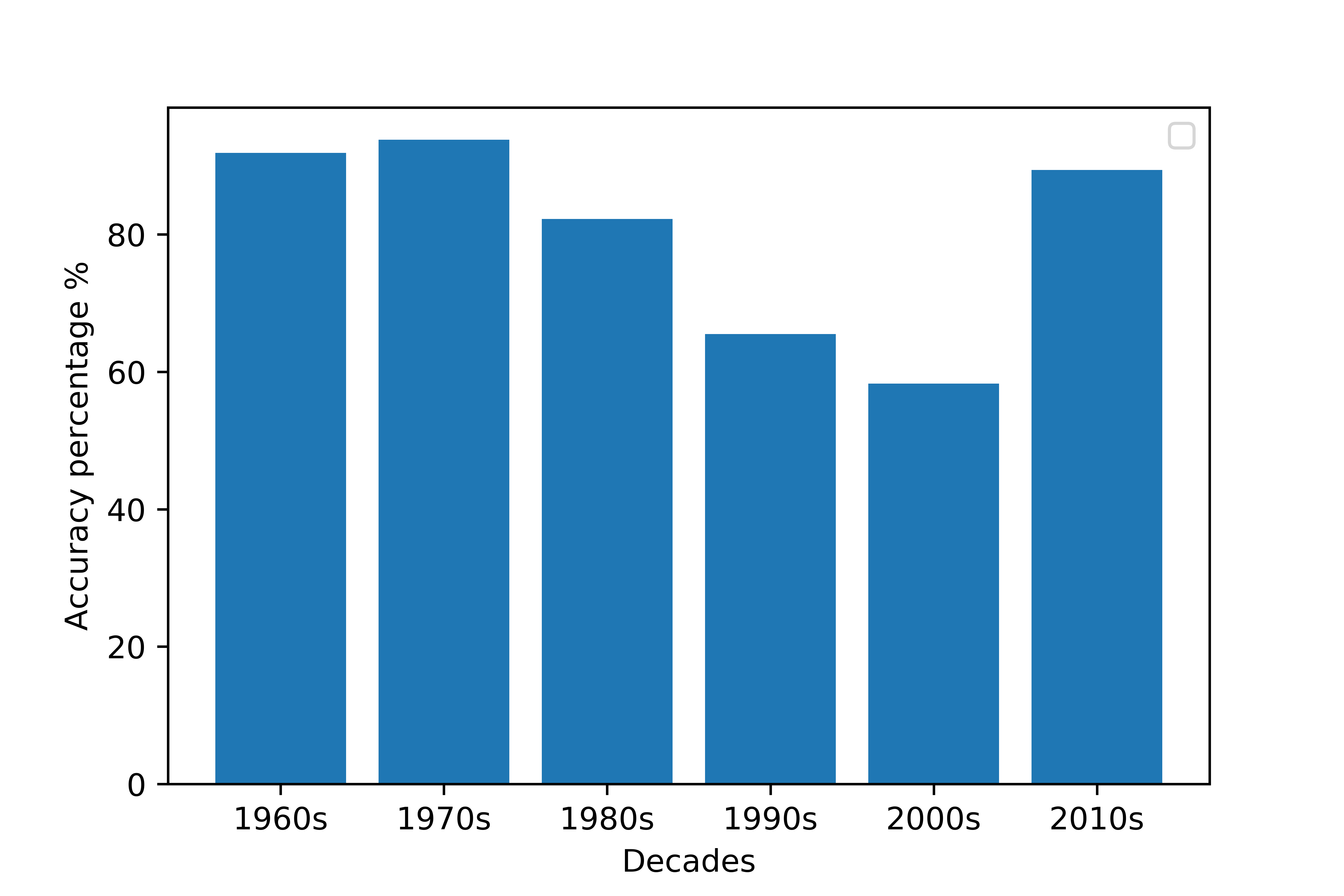}
       \caption{Plot showing accuracy each year}
         \label{fig:trunk_plot}
     \end{subfigure}
     \centering
     \begin{subfigure}[b]{0.44\textwidth}
           \centering
         \includegraphics[width=\textwidth]{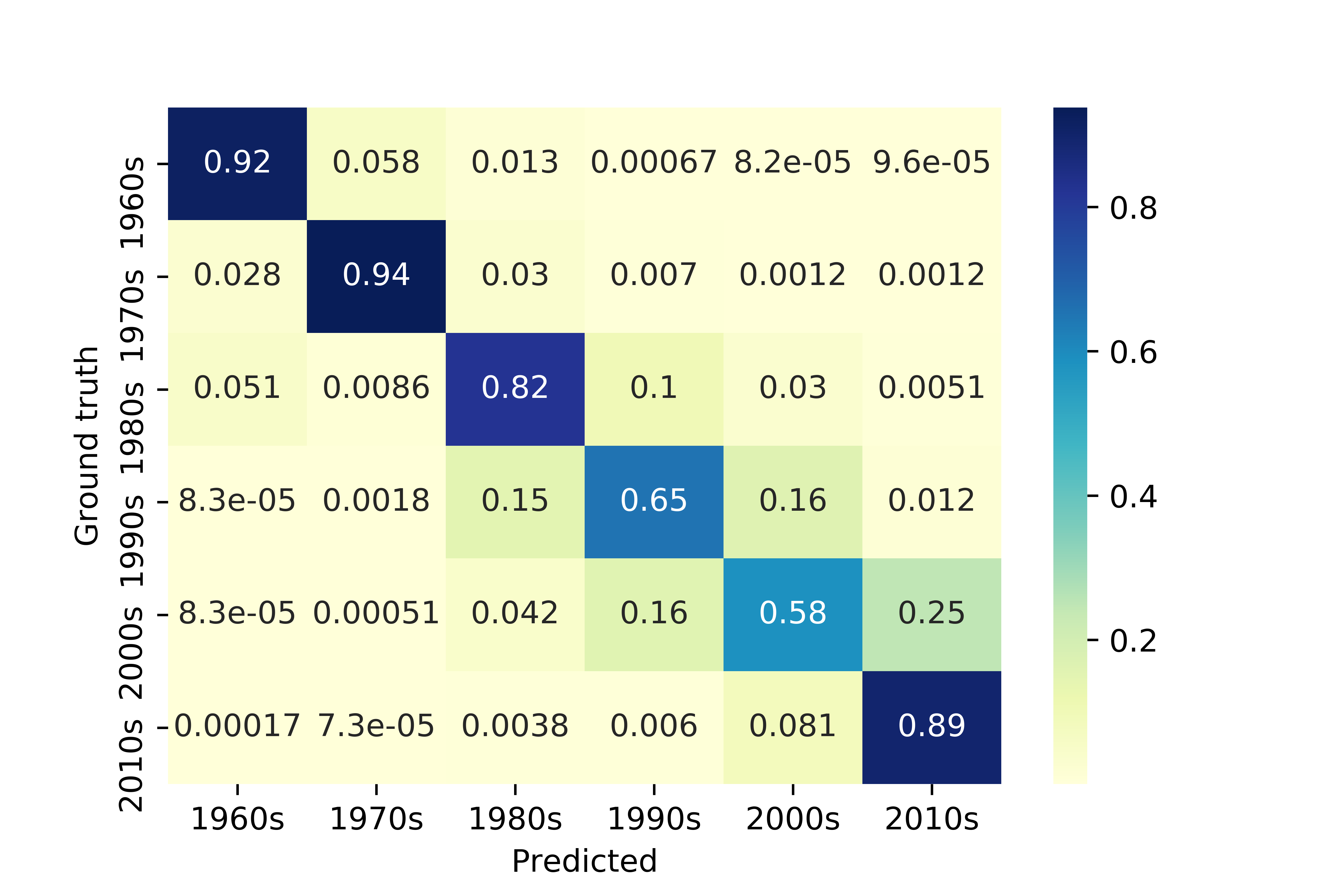}
     \caption{Correlation matrix of truncated data}
         \label{fig:trunk_map}
     \end{subfigure}
      \caption{Plots showing the results of uniform article length}
    \label{fig:trunk}
\end{figure}
 To compare the effect of article length on the model, dataset was processed and all the articles were truncated to have approximately uniform length of words. The model was again trained with the new dataset and tested again. The accuracy of 80\% was observed when it was tested. From the figure \ref{fig:dates_length_original} it can be seen that changing the length does not have an effect on the accuracy decade-wise. Similar pattern as the original dataset was seen here as well. This can be further clearly seen from plot \ref{fig:trunk_plot}  and the correlation map \ref{fig:trunk_map}. With this we are able to prove that the article length did not have an effect on the prediction result.

\subsection{Categorical results}
\begin{figure}[!htb]
   \includegraphics[width=0.5\textwidth]{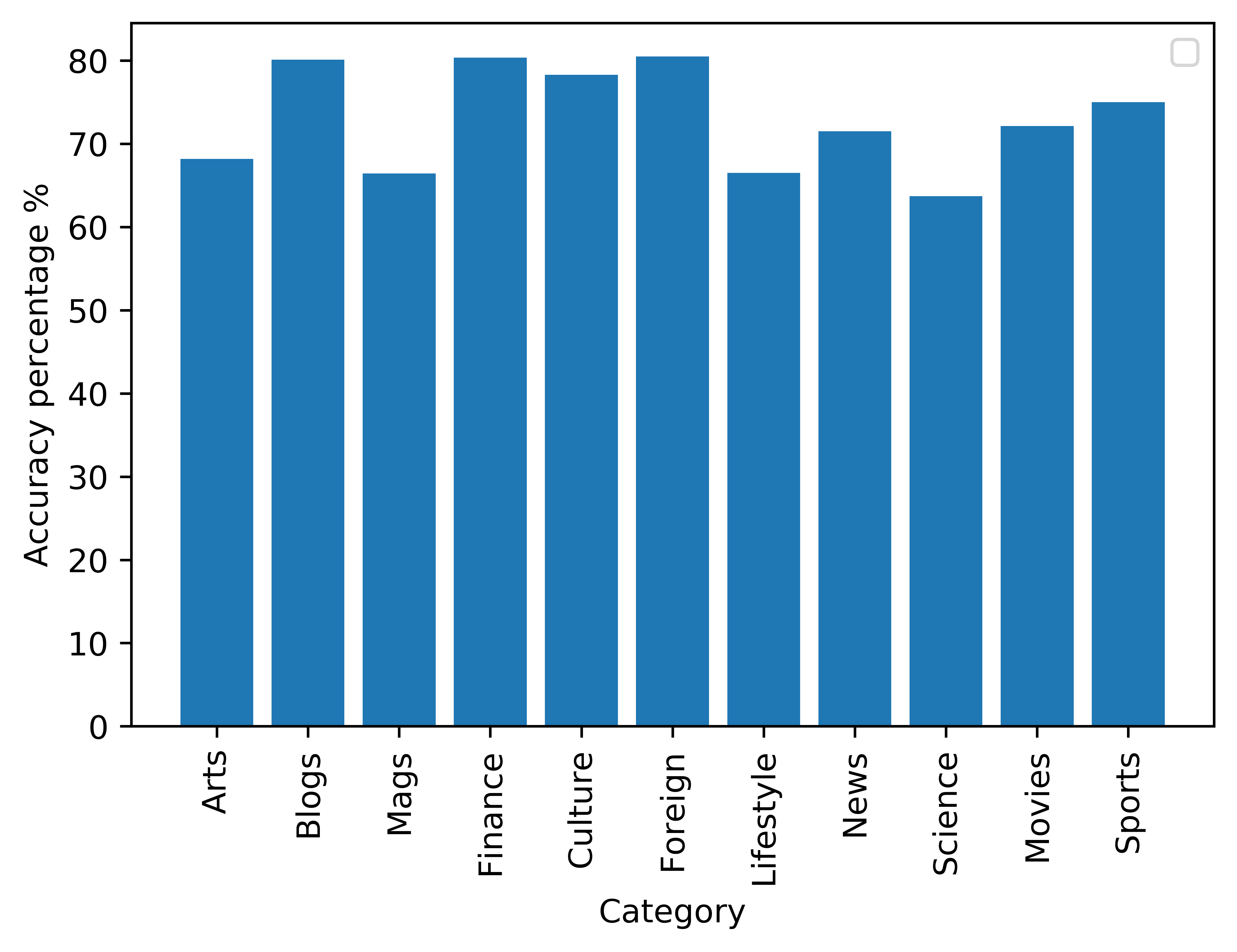}
    \caption{Accuracy Category wise }
    \label{fig:category}
\end{figure}

\begin{figure}[!htb]
   \includegraphics[width=0.5\textwidth]{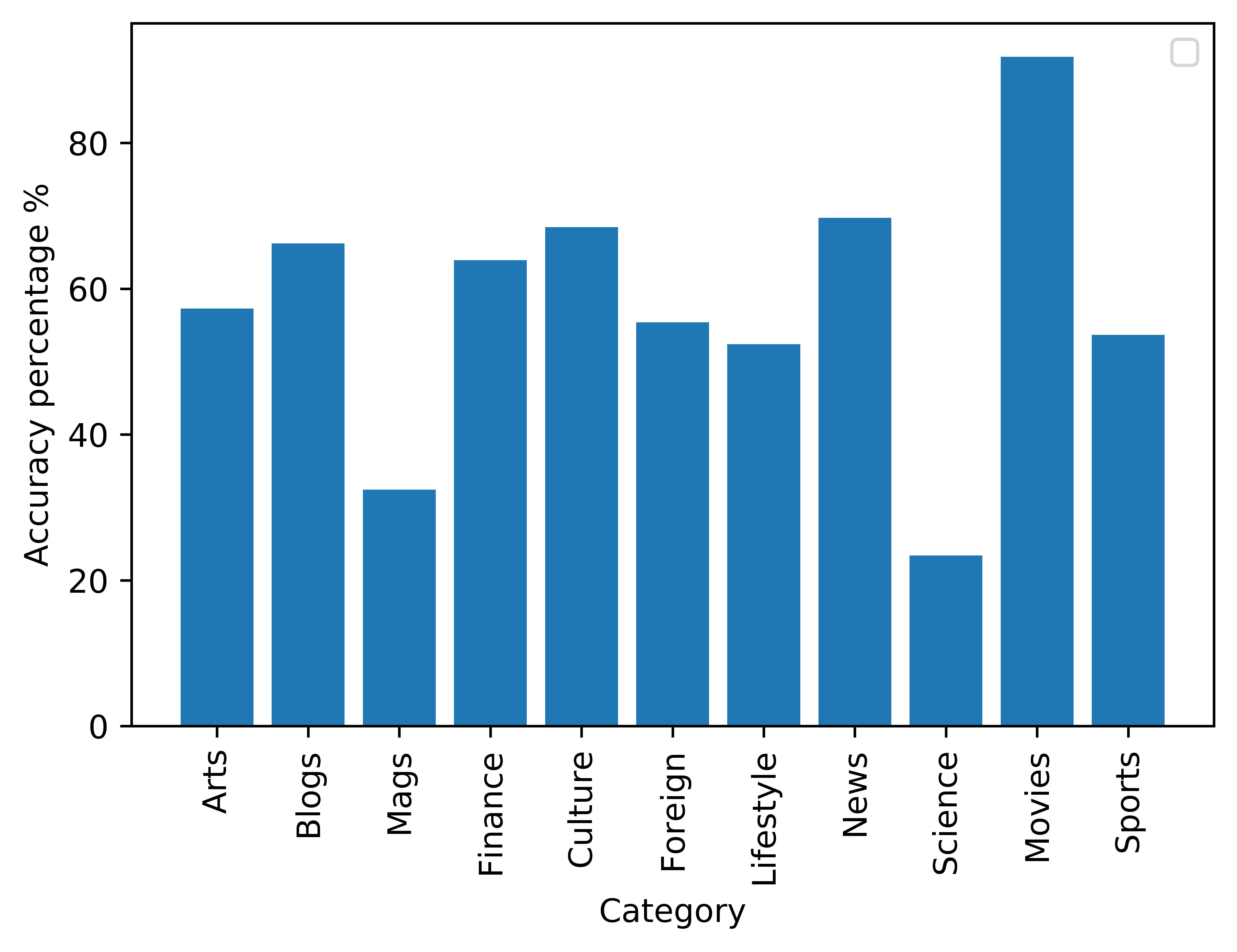}
    \caption{Accuracy of 2000's data across categories }
    \label{fig:2000_category}
\end{figure}

Articles from each category were trained and tested separately. It can be seen that the accuracy varies by the category of the article. The categories such as Finance, Culture,International and Blogs were predicted more accurately than the other categories such as Science,Lifestyle ad Arts. Plot \ref{fig:category} shows the accuracy distribution category wise across all the years. It could be inferred that the words in low-accuracy articles might be changing across decades. A good example will be the Sports accuracy being high, this might be due to the fact that the players change over a period of time and there might be unique things or incidents which would have happened in that particular decade. To check if the accuracy for the 2000s is low across the categories, accuracies were plotted Figure \ref{fig:2000_category}. It could be seen that the movies were pretty distinct having almost 90\%. This could be attributed to the different distinct movies in each decade. Other categories are either bad or average.   

\section{Error Analysis}

Upon analysis, we found that our baseline model and our approach failed in similar articles. This led us to think about what the function of keywords might be in predicting for both models, since our baseline model is purely count-based. One idea is that our models sometimes make the wrong prediction due to the fact that real-world events are not partitioned evenly by decades.

\begin{figure}[!htb]
    \centering
    \textit{9,It is far from clear that Moscow would go along with Mr. Nunn's suggestion since Soviet officials have objected to the idea of confining troop limits to American and Soviet forces. Moscow has proposed that troop limits be set on all NATO and Soviet-bloc forces...}
    \caption{BERT may have thought this article was written in the 1980s due to its topic}
    \label{fig:my_label}
\end{figure}

For example, even though the above article was written in the 1990s, BERT predicted it as being from the 1980s. This could be due to the fact that it has the keywords "Moscow" and "Soviet" several times throughout the paper. Even though the Soviet Union was still very much in the public eye of the United States, keywords relating to those concepts probably popped up much more frequently in articles from the 1980s because that is when the bulk of the cold war happened. 

Another error that we think our models are susceptible to is writing that refers to events that occurred in the past. The source of this error is obvious: if a person uses many keywords that belong to another decade, then BERT will have good reason to predict the decade being written about.

\begin{figure}[!htb]
    \centering
    \textit{1,When I was young I played a dice baseball game for hundreds of hours a year ... I always remembered what Tony La Russa did at the end of the 1988 season. I was 13 staying up late to watch the Los Angeles Dodgers finish La Russa's Athletics ...}
    \caption{Style of writing may not vary enough to denote a decade on its own}
    \label{fig:my_label}
\end{figure}

The above article was written in the current decade, but BERT predicted it as being written in the 1908s. It probably did this because Tony La Russa was the manager of the Red Sox in the 1980s and so it had already seen other articles from the 1980s with that information it them. This leads us to believe that either the size of our model and data is not large enough for style of language to denote a decade on its own or language itself in the United States does not change rapidly enough to predict which decade an article was written in without depending on keyword usage.

\section{Conclusion}

In conclusion, predicting the year of a news article has numerous applications and can provide valuable insights into various fields. In this work, we explored the problem of predicting the publication date of a news article from its text content. We created our own dataset of over 350,000 news articles that were published over six decades from 1960s to 2010s, by scraping the New York Times API. For our baseline model, we implemented a simple Naive Bayes text classifier to classify an article into its respective publication decade, and achieved a baseline accuracy of 63\%. Later, we fine tuned a state of the art pretrained BERT model for our task of text classification, and achieved an accuracy of 82\% which we believe is very impressive. We also analyzed our dataset to show trends that BERT might have picked up in order to reach such a high accuracy, such as the average word length of articles in the dataset increasing with each decade. Finally, we ran our models on articles belonging to specific categories like Sports and Movies, and showed that textual content is a stronger indicator of publication period for some categories like Movies than other categories.

\end{document}